# Correspondence

## Fingerprint Image-Quality Estimation and its Application to Multialgorithm Verification

Hartwig Fronthaler, Klaus Kollreider, Josef Bigun, Julian Fierrez, Fernando Alonso-Fernandez, Javier Ortega-Garcia, and Joaquin Gonzalez-Rodriguez

*Abstract*—Signal-quality awareness has been found to increase recognition rates and to support decisions in multisensor environments significantly. Nevertheless, automatic quality assessment is still an open issue. Here, we study the orientation tensor of fingerprint images to quantify signal impairments, such as noise, lack of structure, blur, with the help of symmetry descriptors. A strongly reduced reference is especially favorable in biometrics, but less information is not sufficient for the approach. This is also supported by numerous experiments involving a simpler quality estimator, a trained method (NFIQ), as well as the human perception of fingerprint quality on several public databases. Furthermore, quality measurements are extensively reused to adapt fusion parameters in a monomodal multialgorithm fingerprint recognition environment. In this study, several trained and nontrained score-level fusion schemes are investigated. A Bayes-based strategy for incorporating experts' past performances and current quality conditions, a novel cascaded scheme for computational efficiency, besides simple fusion rules, is presented. The quantitative results favor quality awareness under all aspects, boosting recognition rates and fusing differently skilled experts efficiently as well as effectively (by training).

*Index Terms*—Adaptive fusion, Bayesian statistics, cascaded fusion, fingerprint, monomodal fusion, quality assessment, structure tensor, symmetry features, training.

## I. INTRODUCTION

AUTOMATIC assessment of image quality by a machine expert is challenging, but useful for a number of tasks: monitoring and adjusting image quality, optimizing algorithms and parameter settings, or benchmarking image-processing systems [1]. Image-quality assessment methods can be divided into full/reduced/no-reference approaches, depending on how much prior information is available on how a perfect candidate image should look like. Here, we study quality assessment of the second kind, where images come from a specific application. General quality metrics originally suggested in image compression studies exist [2] (e.g., mean square error (MSE) or peak signal-to-noise ratio (PSNR)). These earlier approaches are excluded here because of their notorious poor performance in recognition applications, which do not have the same objectives as compression applications.

Manuscript received June 28, 2007; revised January 18, 2008. This work was supported in part by the Spanish Ministry of Education and Science (TEC2006-13141-C03-03), in part by the European Commission (IST-2002-507634), and by the Comunidad de Madrid. The associate editor coordinating the review of this manuscript and approving it for publication was Dr. Anil Jain.

H. Fronthaler, K. Kollreider, and J. Bigun are with Halmstad University, Halmstad SE-30118, Sweden (e-mail: hartwig.fronthaler@ide.hh.se; klaus.kollreider@ide.hh.se; josef.bigun@ide.hh.se).

J. Fierrez, F. Alonso-Fernandez, J. Ortega-Garcia, and J. Gonzalez-Rodriguez are with the ATVS, Escuela Politecnica Superior, Universidad Autonoma de Madrid, Madrid 28049, Spain (e-mail: julian.fierrez@uam.es; fernando.alonso@uam.es; javier.ortega@uam.es; joaquin.gonzalez@uam.es).

In this study, symmetry features [3] are exploited in a local model for generic image quality, applied to fingerprints. We are forced to use models when trying to estimate the quality of biometric images, since a high-quality reference image of the same individual is usually not available (i.e., the link to the individual is not established in advance, for example, by identification). Once available, the benefits of having an automatic image-quality estimate include the following: 1) Assuring quality for all acquired samples and stored templates [4]; 2) adjusting multimodal fusion schemes depending on the quality of the presented samples (e.g., face and fingerprint) [5], [6]; and 3) taking into account the local quality when matching samples [7], [8]. As a result of recent fingerprint verification competitions involving particularly low-quality impressions, even state-of-the-art systems' performance decreases remarkably [9]. Recent advances in fingerprint-quality assessment include [8], [4], [10], and [11]. A taxonomy of fingerprint-quality assessment methods is given in [12]. The novelties of the presented approach will be listed further below.

We are combining fingerprint recognition systems at score level, and refer to it as multialgorithm fusion (in contrast to multimodal fusion). To avoid confusion, we will use "system" or "expert" to address a fingerprint matcher, whereas we refer to a quality assessment method as "method" or "approach." Considering fusion within a modality, in particular, fingerprint recognition [13], [14] showed that combining systems with heterogeneous matching strategies is most desirable, leading to recognition rates that are even higher than when combining the best systems relying on common features. When trying to fuse several experts with unknown skills and matching strategies, some sort of training is advisable to improve the combined performance [15], [16]. The accuracy can be increased even more, if the trained fusion scheme is adaptive as well, meaning that it takes into account current signal conditions trial by trial. This was also confirmed in [5], although unlike here, for a multimodal configuration and employing quality estimates by humans. In [17], the additional information through automatic quality labels was exploited to weight experts, because their individual weaknesses were known *a priori*. Recent studies of fixed and trained fusion strategies include [18] and [19].

This paper improves the state of the art as follows.
- The proposed quality estimation method achieves a continuous modeling of the reference structure. Applied to fingerprints, the benefit is that no misinterpretation of singularities occurs.
- The proposed cascaded fusion scheme is original and saves computation time.
- A trained Bayesian scheme is proven to systematically increase recognition rates of differently skilled experts in quality-adaptive monomodal fusion.

We report quantitative and comparative experimental results of our quality assessment with respect to two existing automatic fingerprint-quality estimation methods [4], [11], and a set of manually assigned quality labels [20], [21]. The QMCYT database and two databases of the FVC2004 [9] were employed in this evaluation. Additionally, three fingerprint recognition systems [4], [7], [20] are used to 1) benchmark the quality labels and 2) carry out the quality-adaptive multialgorithm fusion.

## II. QUALITY ESTIMATION

In the first part of this section, a more general description of the suggested automatic quality assessment method is given. The ideas are then adapted to fingerprint-quality estimation. Its applicability to other biometric modalities was indicated by means of face images in [22].

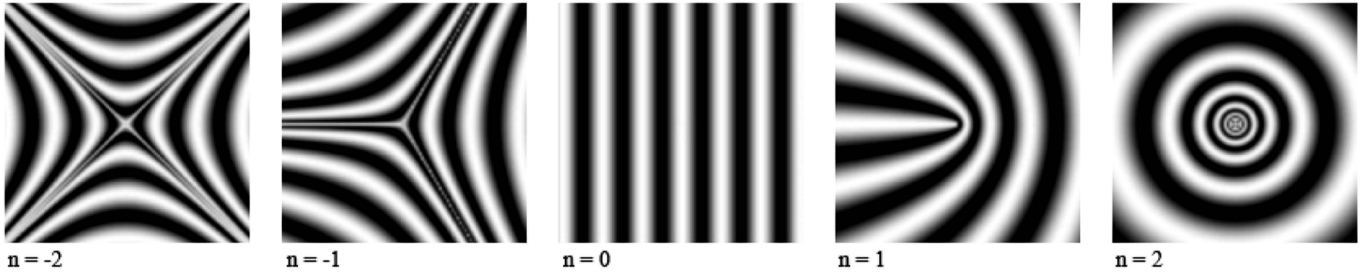

Fig. 1. Patterns with orientation description $z = \exp(\text{in}\phi)$: Straight lines for $n = 0$ (linear symmetry); parabolic curves and line endings for $n = \pm 1$ (parabolic symmetry); stars, circles, and spirals for $n = \pm 2$ (circular symmetry).

### A. Quality Assessment Features

The orientation tensor holds edge and texture information, which is exploited in this study to assess the quality of an image. We wish to determine whether this information is structured and generic in some sense [i.e., to distinguish noisy content from relevant nontrivial structures) (Fig. 1)]. The latter are among others essential for many recognition algorithms, representing the individuality of a biometric signal, but also have significance in low-level human vision models enabling object recognition and tracking [23]. Our method decomposes the orientation tensor of an image into symmetry representations, where the included symmetries are related to the particular definition of quality and encode the *a priori* content knowledge about the application (e.g., fingerprints and face images). The resulting quality metric mirrors how well a test image comprises the expected symmetries.

The orientation tensor is given by the equation

$$z = (D_x f + i D_y f)^2 \qquad (1)$$

where $D_x f$ and $D_y f$ denote the partial derivatives of the image with respect to the $x$ and $y$ axes. The squared complex notation directly encodes the double angle representation [3]. For the computation of the derivatives, separable Gaussians with a small standard deviation $\sigma_1$ are used. Next, the orientation tensor is decomposed into symmetry features of order $n$, where the $n$th symmetry is given by $\exp(\text{in}\phi + \alpha)$ [3], [23]–[25] representing the argument of (1). The corresponding patterns are shown in Fig. 1 (e.g., straight lines for $n = 0$, parabolic curves, and line endings for $n = \pm 1$). Higher orders include circular, spiral, and star patterns. In Fig. 1, the so-called class member $\alpha$, which represents the global orientation of the pattern, is zero. Filters modeling these symmetry descriptions can be obtained by

$$h_n = (x + iy)^n \cdot g, \qquad \text{for} \quad n \geq 0 \qquad (2a)$$
$$h_n = (x - iy)^{|n|} \cdot g, \qquad \text{for} \quad n < 0 \qquad (2b)$$

where $g$ denotes a 2-D Gaussian with standard deviation $\sigma_2$ in the $x$ and $y$ direction. These features are algebraic invariants of physical operations [e.g., translation, rotation, and zooming (locally)]. For a more detailed review of symmetry filters and the symmetry derivatives of Gaussians, we refer to [3]. Decomposing an image into certain symmetries involves calculating $\langle z, h_n \rangle$, where $\langle \cdot, \cdot \rangle$ denotes the 2-D scalar product, yielding complex responses $s_n = c \cdot \exp(i\alpha)$, with $c$ representing the certainty of occurrence and $\alpha$ (class member) encodes the direction of symmetry $n$ (for $n \neq 2$). Normalized filter responses are obtained by calculating

$$s_n = \frac{\langle z, h_n \rangle}{\langle |z|, h_0 \rangle} \qquad (3)$$

where the nominator is the total energy of the symmetry (all possible orders) [3]. In this way, $\{s_n\}_{n \in N}$ describe the symmetry properties of an image in terms of $|N|$ orders. The definition of quality for a specific application determines the expected orders ($N$) and scales ($\sigma$) used for the reference model. Furthermore, we demand $\{s_n\}$ to be well separated over the image plane, in which we look for a high and dominant symmetry at each point. Equation (4) denotes an inhibition scheme [23]

$$s_n^I = s_n \cdot \prod_{k \in N \setminus n} (1 - |s_k|) \qquad (4)$$

where $k$ refers to the remaining applied orders, to sharpen the spatial extension of filter responses, and I is a label that stands for inhibition. Consequently, a high certainty of one symmetry type requires a reduction of the other types. We calculate the covariance among $\{|s_n^I|\}$ in blocks of size $b \times b$ in order to test whether the filter responses have been mutually exclusive. A large negative covariance supports that this is the case, and the neighborhood behaves as a high-quality local image. On the other hand, positive covariance implies the co-occurrence of mutually exclusive symmetry types in the vicinity of a point, which is an indication of noise or blur. We incorporate this information by weighting the symmetry certainty. We sum $\{|s_n^I|\}$ over $n$ at each pixel, resulting in a total symmetry image

$$s = \sum_{n \in N} |s_n^I|. \qquad (5)$$

Image $s$ is further averaged within blocks (tiles) of size $b \times b$, yielding $\bar{s}$ (we use $\bar{\phantom{x}}$ to denote block-wise operating variables). The quality measure $\bar{q}$ for each block is then computed as follows:

$$\bar{q} = m(\bar{r}) \cdot \bar{s} \qquad (6)$$

where $\bar{r}$ denotes the block-wise correlation coefficient, and $m$ is a monotonically decreasing function, so that $m : [-1, 1] \to [0, 1]$. The quantity $\bar{r}$ is calculated as an average of the correlation coefficients among $\{|s_n^I|\}_{n \in N}$, that is, between any two involved orders $\bar{r}_{k,l}$, as defined by

$$\bar{r}_{k,l} = \frac{\text{Cov}\left(|s_k^I|, |s_l^I|\right)}{\sqrt{\text{Var}(|s_k^I|)\text{Var}(|s_l^I|)}}. \qquad (7)$$

Note that $\bar{r}_{k,l} = \bar{r}_{l,k}$, and that in case of employing only two orders for the decomposition (e.g., $N = \{0, 1\}$, $\bar{r}$ is equal to $\bar{r}_{01}$). An overall quality metric is established by averaging $\bar{q}$ over the "interesting" blocks $\bar{i}$, which are represented by blocks where $\bar{s} > \tau_s$, thus having a minimum total symmetry response. The proposed technique is implemented and tested by means of automatic fingerprint image-quality estimation.

*B. Fingerprint-Quality Estimation*

By human opinion, the quality of a fingerprint image is usually expressed in terms of the clarity of ridge and valley structures, as well as the extractability of certain points (minutiae, singular points) [8]. In our approach, we concentrate on medium-to-global-scale features of a fingerprint, represented by the orientation, singular points, scratches, and low-contrast areas. The purpose is to identify and grade "bad" blocks, so that any subsequent analysis of the fingerprint is alleviated. It is, for example, a main problem for minutiae detection methods to distinguish genuine minutia points from similar patterns stemming from scratches. However, in our approach, these neighborhoods will already be marked because we act on a higher level and shall detect only the scratches. Another important point is to include both highly and lowly curved structures in the quality definition, because otherwise it cannot model the global ridge-valley flow. We employ large filters for two symmetry types $n = 0$ and $n = 1$. The former is known to model the typical ridge-valley flow well, whereas the latter has been shown to model the flow about the singular regions of a fingerprint (compare Fig. 1) [7], [26]. Intuitively, features of order $|n| > 1$ are not considered meaningful here. Now, in noisy, low-contrast regions both symmetries are present which contradicts our quality definition (positive correlation). Likewise, both symmetries are low along a scratch. In singular regions, parabolic symmetry ($n = 1$) dominates clearly while linear symmetry ($n = 0$) does so in the remaining area. Given a reasonably good quality fingerprint, this yields alternating, very high and low symmetries, and a negative correlation, fitting our quality definition. Only three scalar products are needed with the orientation tensor $\langle z, h_0 \rangle$, $\langle |z|, h_0 \rangle$, and $\langle z, h_1 \rangle$. Implemented by means of 1-D convolutions with Gaussian (derivative) filters on initially downsized images, the approach is also fast. More precisely, the space and time complexity is $O(w^2 \cdot n)$, where $w$ and $n$ are the largest image dimension and the length of the 1-D filter, respectively. The linear function $\frac{1}{2}(1 - \bar{r})$ is employed for $m$ in (6). The used value for block size $b$ is eight pixels, the symmetry features use a $\sigma_2$ of 3. For the construction of the orientation tensor, a $\sigma_1$ of 0.6 is employed. These values were chosen in an optimization search, and small variations will not affect the functionality of the method.

Fig. 2 depicts decompositions of two example fingerprints of the QMCYT database into $s_0^I$ (linear symmetry) and $s_1^I$ (parabolic symmetry). The final column shows the combined symmetry $s$, in which bright areas indicate well-defined ridge-valley structure in both low and high curvature regions. The quality of the first fingerprint is rather high, apart from some scars. The latter can be traced through the decomposition. In the second row in Fig. 2, we show an impression that suggests dry skin conditions, which affect the quality. Also in this case, clear fingerprint structure results in bright areas in the respective symmetry and in $s$, whereas structural absence is reflected by the darker regions. The tiled images representing the block-wise variables for the example fingerprints are displayed in Fig. 3. The block-wise average of $s$ is represented by $\bar{s}$ in the first column. Fingerprint segmentation is done implicitly via $\bar{i}$, as shown in the second pair of images. In the third column, we observe that the covariance is negative in reasonably good-quality regions, whereas it is positive in noisy and low-contrast regions (i.e., it detects scratches and imperfections). This separation is not so apparent when considering $\bar{s}$ only. The final block-wise quality $\bar{q}$ is depicted by the last pair of images.

Previous methods for (local) fingerprint-quality assessment have been exploiting the spatial coherence of the ridge flow only, by essentially determining or approximating $s_0$ [12]. Additionally, the latter has commonly been partitioned into blocks $\bar{s}_0$. Inspecting Fig. 4 reveals that this strategy may not be enough, because important regions, such as singular points (e.g., core, delta) are per definition incoherent to the ridge flow, and their strong presence therefore automatically impairs the estimated quality. Focusing on the second row, we see how severely the single core and two delta points distort the quality map $\bar{s}_0$. Note the different shape of the singular point regions not leading to different results for $\bar{q}$, though. This is due to the $h_1$-filter's response to both prominent singular point types "core" ($n = 1$) and "delta" ($n = -1$), because the former is implicitly contained in subpatterns of the latter. Therefore, when estimating an overall quality metric by averaging the quality map, $\bar{q}$ is expected to be more suitable than $\bar{s}_0$. Quantitative results with comparisons will be presented further. To our best knowledge, there is no other reported method that measures the quality of a typical and high curvature ridge-valley structure.

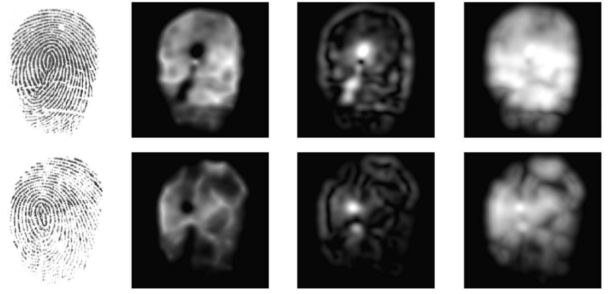

Fig. 2. Decomposition of sample fingerprints. C1: Original fingerprint. C2: linear symmetry magnitude $|s_0^I|$. C3: parabolic symmetry magnitude $|s_1^I|$. C4: "total symmetry" (summed magnitudes) contains relevant portions $|s|$.

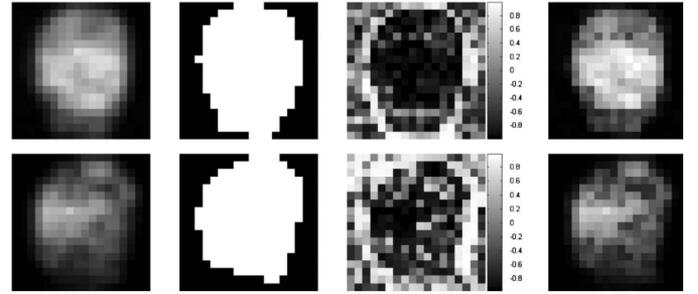

Fig. 3. Fingerprint-quality estimation. C1: Block-wise averaged total symmetry $\bar{s}$. C2: Thresholded total symmetry $\bar{i}$. C3: Correlation coefficient between the parabolic and linear symmetry $\bar{r}$. C4: Tiled quality measure ($\bar{q}$).

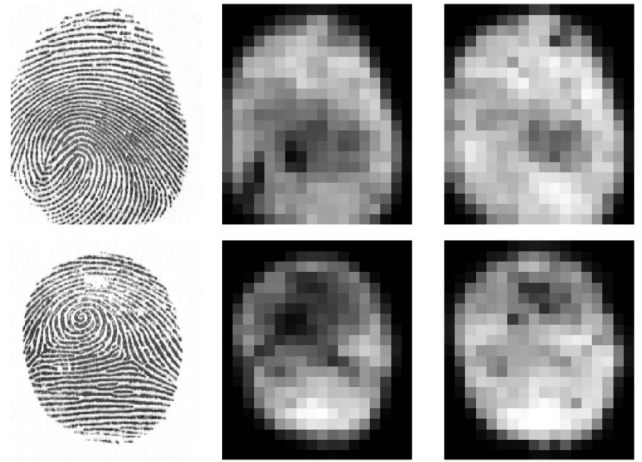

Fig. 4. Illustrating the difference $\bar{s}_0$ (b) and $\bar{q}$ (c): Here we can see that the singular points are misinterpreted in terms of quality when just averaging $s_0$.

## III. FUSION

In this section, we will derive different multialgorithm fusion schemes. The quality-adaptive strategies weight several recognition experts according to their confidence measures. This is done in a continuous way in a Bayes-based training fusion (Section III-A), and in a more aggressive fashion in a cascaded type of fusion (Section III-B). Confidence measures are modeled by the fingerprints' overall quality in both cases. A listing of simple (nonadaptive) fusion schemes closes the section.

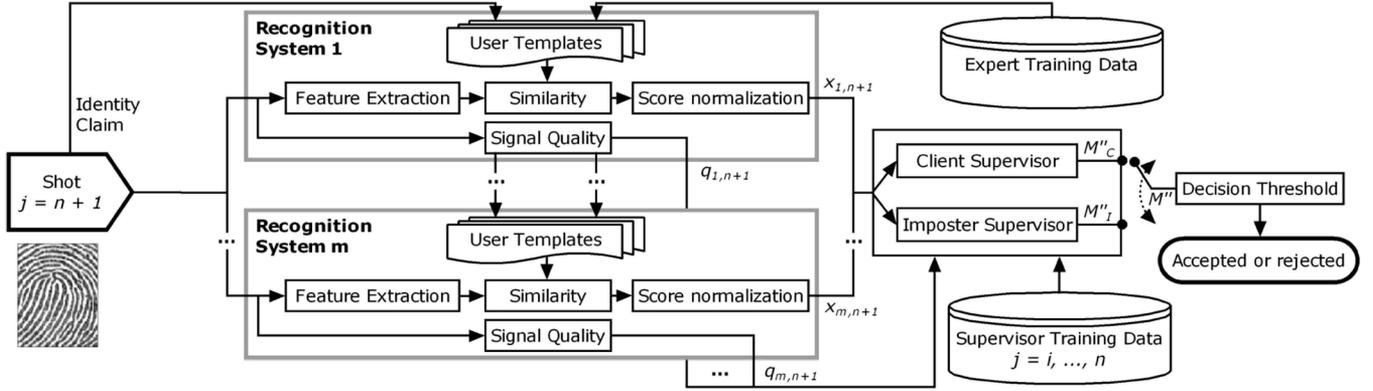

Fig. 5. Multialgorithm system model: Schematics including all components of the proposed Bayesian supervisor. All experts deliver a certainty in addition to their score, which is estimated as the image quality here.

### A. Bayesian Supervisor

This section is devoted to an adaptive fusion scheme using Bayes theory [27]. For a more profound description of the employed model, we refer to [15]. Its probabilistic background is further detailed in [28] and [29]. As indicated in Fig. 5, we combine independent fingerprint recognition systems yielding a monomodal multialgorithm environment. An input fingerprint is referred to as a shot. For every shot, we have several different experts' opinions delivered to the Bayesian supervisor. The following notation is used when describing the statistical model and the supervisor within this paper:

$i$     index of the experts $i \in 1 \ldots m$;
$j$     index of shots $j \in 1 \ldots n, n+1$;
$x_{ij}$     authenticity score computed by expert $i$ based on shot $j$;
$s_{ij}$     variance of $x_{ij}$ (estimated by expert $i$);
$y_j$     true authenticity score of shot $j$;
$z_{ij}$     error (misidentification) score of an expert $z_{ij} = y_j - x_{ij}$.

The true authenticity score $y_j$ can only take two numerical values, namely "True" or "False." So if the values of $x_{ij}$ are between 0 and 1, the values of $y_j$ are chosen to be 0 and 1, respectively. The training of the supervisor is performed on the shots $j \in 1 \ldots n$, where $x_{ij}$ and $y_j$ are known. When the supervisor is operational, we consider the shot $j = n+1$ as a test shot. In this case only, $x_{i,n+1}$ is known and the task of the supervisor is to estimate $y_{i,n+1}$. It is assumed that the single experts and the supervisor are trained on different sets. Note that the experts provide a quality estimate in addition to each score which is modeled to be inversely proportional to $s_{ij}$. This variance is then used by the supervisor for evaluation.

*1) Statistical Model:* The employed adaptive fusion strategy uses Bayesian statistics and assumes the errors of the single experts to be normally distributed (i.e., $z_{ij}$ is considered to be a sample of the random variable $Z_{ij} \sim N(b_i, \sigma_{ij}^2)$). This does not strictly hold for common audio- and video-based biometric machine experts [15]. Nevertheless, it was shown that this problem can be addressed by considering client and impostor distributions separately. Thus, the following two supervisors representing the expert opinions $y_j = 1$ and $y_j = 0$ are constructed:

$$\mathcal{C} = \{x_{ij}, s_{ij} | y_j = 1 \text{ and } 1 \leq j \leq n\} \quad (8)$$
$$\mathcal{I} = \{x_{ij}, s_{ij} | y_j = 0 \text{ and } 1 \leq j \leq n\}. \quad (9)$$

The two supervisors will be referred to as client supervisor and impostor supervisor, respectively.

The task of the client supervisor is to estimate the expected true authenticity score $y_j$ based on its knowledge of client data (i.e., computing $M_{\mathcal{C}}'' = E[Y_{n+1} | \mathcal{C}, x_{i,n+1}]$). The prime notation is used to distinguish the three different supervisor states. No prime means training, one denotes calibration, and two indicate the authentication (operational) phase. The impostor supervisor estimates $y_j$ by computing $M_{\mathcal{I}}'' = E[Y_{n+1} | \mathcal{I}, x_{i,n+1}]$.

The supervisor, which comes closer to the ideal case (1 for the client supervisor, 0 for the impostor supervisor), is considered as the final conciliated overall score $M''$

$$M'' = \begin{cases} M_{\mathcal{C}}'', & \text{if } |1 - M_{\mathcal{C}}''| - |0 - M_{\mathcal{I}}''| < 0 \\ M_{\mathcal{I}}'', & \text{otherwise.} \end{cases} \quad (10)$$

*2) Supervisor:* Having the experts scores and the quality estimates, the Bayesian supervisor can be summarized as follows.
1) Training phase: In case of the client supervisor, the bias parameters for all experts are estimated as follows:

$$M_{\mathcal{C}i} = \frac{\sum_j \frac{z_{ij}}{\bar{\sigma}_{ij}^2}}{\sum_j \frac{1}{\bar{\sigma}_{ij}^2}} \quad \text{and} \quad V_{\mathcal{C}i} = \frac{1}{\sum_j \frac{1}{\bar{\sigma}_{ij}^2}} \quad (11)$$

here, $j$ is the index of the training set $\mathcal{C}$. The variances $\bar{\sigma}_{ij}^2$ are calculated by $\bar{\sigma}_{ij}^2 = s_{ij} \cdot \alpha_{\mathcal{C}i}$, where

$$\alpha_{\mathcal{C}i} = \frac{\left( \sum_j \frac{z_{ij}^2}{s_{ij}} - \left( \sum_j \frac{z_{ij}}{s_{ij}} \right)^2 \left( \sum_j \frac{1}{s_{ij}} \right)^{-1} \right)}{n_{\mathcal{C}} - 3} \quad (12)$$

where $n_{\mathcal{C}}$ denotes the number of shots in $\mathcal{C}$. If one or more experts do not provide any quality estimates, $s_{ij}$ is set to 1. The bias parameters $M_{\mathcal{I}i}$ and $V_{\mathcal{I}i}$ for the impostor supervisor can be estimated similarly.

2) Operational phase: At this stage, authentication on "live" data is performed (i.e., the time instant is $n+1$ and the trained supervisors can access the expert opinions $x_{i,n+1}$ but not the true authenticity score $y_{n+1}$). In a first step, the client and impostor supervisors have to be calibrated regarding to their past performance. In case of the client supervisor, this calibration is denoted by

$$M'_{\mathcal{C}i} = x_{i,n+1} + M_{\mathcal{C}i} \quad \text{and} \quad V'_{\mathcal{C}i} = s_{i,n+1} \cdot \alpha_{\mathcal{C}i} + V_{\mathcal{C}i}. \quad (13)$$

Having the calibrated experts, they are combined as follows:

$$M''_{\mathcal{C}} = \frac{\sum_{i=1}^{m} \frac{M'_{\mathcal{C}i}}{V'_{\mathcal{C}i}}}{\sum_{i=1}^{m} \frac{1}{V'_{\mathcal{C}i}}}. \quad (14)$$

The computations for the impostor case ($M'_{\mathcal{I}}, V'_{\mathcal{I}}$, and $M''_{\mathcal{I}}$) follow the same pattern. The final supervisor decision is made according to (10).

*3) Quality Adaptive Strategy:* As indicated in Fig. 5, each expert provides a score $x_{ij}$ and a quality estimate $q_{ij}$ for every single authentication assessment. The quality measure is not an estimation of the general reliability of the expert itself. It is considered to be a certainty measure for the current score based on the quality of the input shot. So we propose to calculate $s_{ij}$ using the qualitative knowledge of the experts on the input biometric data they assess. Section II details our approach to extract such a quality estimate from a shot. In the second part of (13), the trained supervisor adapts the weights of the experts employing the input signal quality. We define quality index $q_{ij}$ of the score $x_{ij}$ as follows:

$$q_{ij} = \min(Q_{ij}, Q_{i,\text{claim}}) \quad (15)$$

where $Q_{ij}$ is the quality estimate produced by expert $i$ in shot $j$ and $Q_{i,\text{claim}}$ is the average quality of the biometric samples used by expert $i$ for modeling the claimed identity. All quality values are in the range $[0, q_{\max}]$ where $q_{\max} > 1$. In this scale, 0 is the poorest quality, 1 is considered as normal quality, and $q_{\max}$ corresponds to the highest quality. The final variance parameter $s_{ij}$ of the score $x_{ij}$ is obtained by

$$s_{ij} = \frac{1}{q_{ij}^2}. \quad (16)$$

Training is the key point of the Bayes-based fusion approach. The biases $M_{\mathcal{C}i}/M_{\mathcal{I}i}$ and $V_{\mathcal{C}i}/V_{\mathcal{I}i}$ of expert $i$ evaluated during training are used to weight the experts' scores in the joint accept/reject decision. This is done in nonadaptive fusion without considering any experts' confidences into their scores. In adaptive fusion, these confidences are included with $s_{ij} \neq 1$ to the effect that low confidence in its score for the current claim decreases the expert's say in the joint decision. Since the confidences are modeled by signal qualities, a dependency between quality and the expert's recognition performance has been estimated during training. This is exploited in the operational phase to continuously shift decision power among experts. The usage of the procedure described before in multialgorithm fusion as well as with automatically derived quality signals is novel.

*B. Cascaded Fusion*

One can argue that the computation time is problematic if several systems have to be executed for every single match (i.e., for identification within a large database, for example, US-Visit). A reasonable way to address this issue is to dynamically include further experts if a single one cannot come up with a clear decision. In such a configuration, a minimal number of experts is active most of the time, while still getting the benefits of fusion (improved recognition rates). This is also visualized in Fig. 6, where we see a series of systems—primary, secondary, etc. systems in the following—triggered by certainty thresholds, meaning that system $i$ is utilized if and only if $c_{i-1}$ is below a certain threshold. Afterwards, all available scores $x_i$ are fused according to a fusion rule $f$, which can be chosen simple. This configuration is inspired by cascaded classifiers [30] (i.e., degenerate decision trees [31]). Using scores themselves as certainty thresholds is not recommendable since they are naturally low in most of the cases for identification, and they might be wrong as well. In contrast, image quality is practicable, since the probability of a false acceptance or rejection is higher if the quality of the involved impressions is lower, while fusion should essentially oppose this fact. The image quality used as certainty threshold is relatively independent of the single experts, such that $c_i$ can be shortened to $c$ (compare Fig. 6). So the number of experts included into the current decision is determined by a single certainty. A trickier question is how to decide on the "trigger" thresholds $\tau_1, \ldots, \tau_{m-1}$. Intuitively, one chooses $\tau_1 > \tau_i > \tau_{m-1}$, since more experts shall be utilized with decreasing signal quality. We suggest setting $\tau_1$ to half the expected best fingerprint quality, $\tau_2$ to half of the remaining quality interval, etc., such that $\tau_i = 0.5 \cdot \tau_{i-1}$. Assuming a uniform distribution of the fingerprints' quality, the number of expert executions for $m$ cascaded systems is expected to be $\sum_{i=0}^{m-1} \frac{N}{2^i}$, where $N$ is the total number of trials (the primary system has to be executed $N$ times). This yields $\frac{2-2^{1-m}}{m} \cdot 100$ percent expert executions. As to the expected error rate, we cannot easily derive a similar prediction, because it depends on the employed fusion rule as well as on the ordering of systems. Being an initial study of the novel fusion scheme, we do not formalize this here. However, no loss of recognition accuracy should be possible for certain thresholds, and reasonable loss is expected for the ones already suggested. While this is a guideline, we will reflect its applicability when we find optimal thresholds by a systematic search in the next section. It would be further desirable if the quality assessment method and the primary system shared computational steps to save resources.

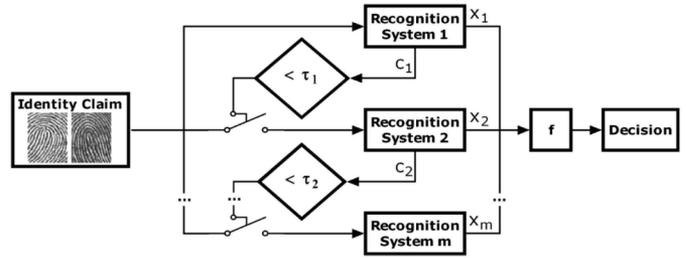

Fig. 6. Cascaded fusion: Experts are triggered on demand and combined only under uncertainty (here: bad quality).

*C. Simple Schemes*

Past experiments indicated that combining systems in simple ways could already lead to relatively good results. Such fusion schemes include, for example, SUM and MAX rules, meaning that the average, respectively, of the maximum of all experts' scores is taken as the final score. Since they are nonadaptive, we also refer to them as global MAX, global SUM, etc. It has been claimed in several studies that simple schemes are not clearly outperformed by trained (nonadaptive) strategies, for example, support vector machines, in neither monomodal fusion [14] nor multimodal fusion [18]. Simple, yet adaptive schemes have been successfully applied in quality-based multialgorithm fusion [17]. In our study, only nonadaptive simple schemes are used to facilitate comparison.

IV. EXPERIMENTS

An approach to measure the impact of signal quality on the recognition performance is to divide the database into several quality groups and to run recognition tests within them. Inversely, given a correct quality division, one expects monotonously decreasing error rates for groups of increasing quality. To benchmark the proposed quality assessment method, we compare it to 1) human grading; 2) National Institute of Standards and Technology Fingerprint Image Quality (NFIQ) [4]; and 3) the local orientation quality score (LOQ) [11]. The latter analyses a fingerprint's quality in blocks by computing the average absolute difference in orientation angle between the surrounding blocks. A smooth change in orientation is interpreted as high quality. It is therefore clear that singular points, where the orientation changes per definition abruptly, are downgraded, which is unfavorable as elaborated in Section II-B. NFIQ is an intensely trained quality assessment method, which is part of NIST FIS2[1] [32]. The NFIQ implementation is based on 5244 impressions for training.

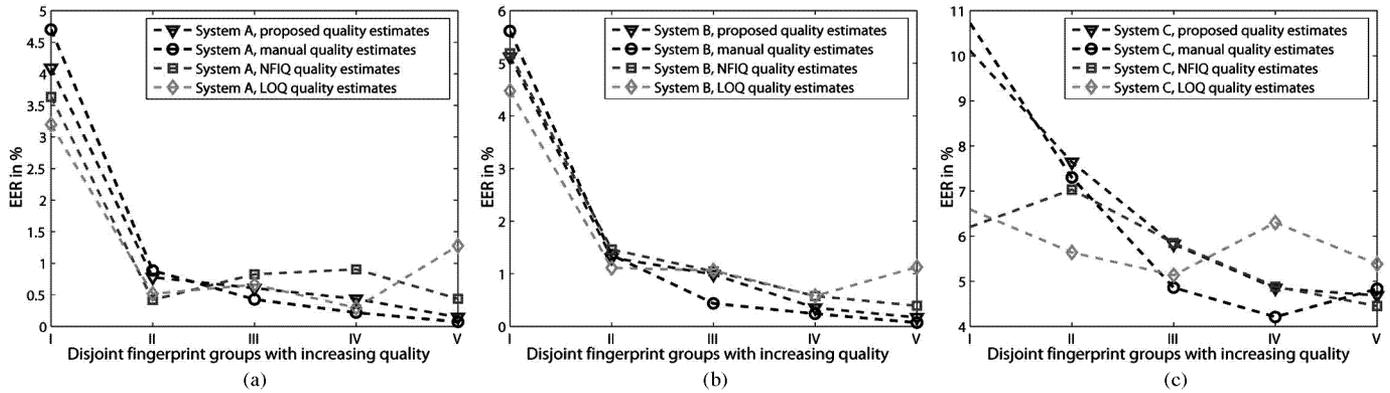

Fig. 7. EER for systems A, B, and C (from left to right) within quality groups I-V from the QMCYT database. The partitions are established by means of different quality assessment methods (see legend).

In this study, all experiments are conducted on the QMCYT fingerprint database [21], and some on two databases employed in FVC2004 [9]. The former defines $75 \times 10$ fingerprints $\times 12$ impressions, whereas the latter contain 100 fingerprints $\times 8$ impressions per database. For each impression in the QMCYT database, a manually annotated quality label is available [21]. We employ a recently developed fingerprint recognition system [7], called system A in the following to validate the quality estimates. To investigate feature independency, we also employ the NIST FIS2—referred to as system B—in a similar test. Note that system B is entirely minutiae based whereas system A exploits both minutia and texture features for fingerprint alignment and matching, respectively. As a third expert, system C represents a nonminutia-based recognition system utilizing Gabor features, as described in [20]. The 750 fingerprints of the QMCYT database are split into five equally sized partitions of increasing quality. The criterion for a fingerprint to be part of a certain group I-V is the average quality index for its genuine trials (impressions). The latter are chosen to be $150 \times 9$ per group, while $150 \times 74$ impostor trials are performed, considering fingers of the same type only as impostors (one impression). We show the EER of system A, B, and C for all quality groups, which have been established according to the different quality assessment methods (see Fig. 7). According to the EER curves, we can observe that the proposed method shows most similar behavior to the manual estimates (human opinion). It is worth mentioning that the grading by the proposed method and LOQ is continuous in $[0\ldots 1]$, whereas it is discrete for NFIQ and the human opinion being in $[1\ldots 5]$ and $[0\ldots 9]$, respectively. The latter two output ranges are normalized into $[0\ldots 1]$. The same experiment is repeated for databases DB2 and DB3 employed in FVC2004. The 100 fingerprints of each database are split into partitions following the rules from before. For each database and per quality group, $20 \times 28$ genuine trials and $20 \times 99$ impostor trials are performed. We show the EER of systems A and B for all quality groups in the top row (DB2) and bottom row (DB3) in Fig. 8. System C and LOQ are left out due to the undesirable findings in the previous experiment. When looking at Fig. 8, we can observe a generally higher EER level and variance. The correct estimation of the different quality categories has more of an impact on recognition rates (compare Fig. 7) due to the increased difficulty of the FVC2004 databases. The severe image-quality impairments were obviously detected well by both quality estimators. In particular, the proposed method leads to monotonically decreasing EER curves for all involved recognition systems and databases. This strengthens our claim that including all fingerprint regions in the assessment yields the most reliable quality labels. Furthermore, the results confirm the usefulness of the employed symmetry features and their energy-independent usage in our algorithm (using normalized filter answers), without especially adapting it to the different databases. In Table I, we state the EER for each recognition system (A, B, and C) over the whole QMCYT database (i.e., when the quality division is dissolved again).

[1]NIST Fingerprint Image Software 2

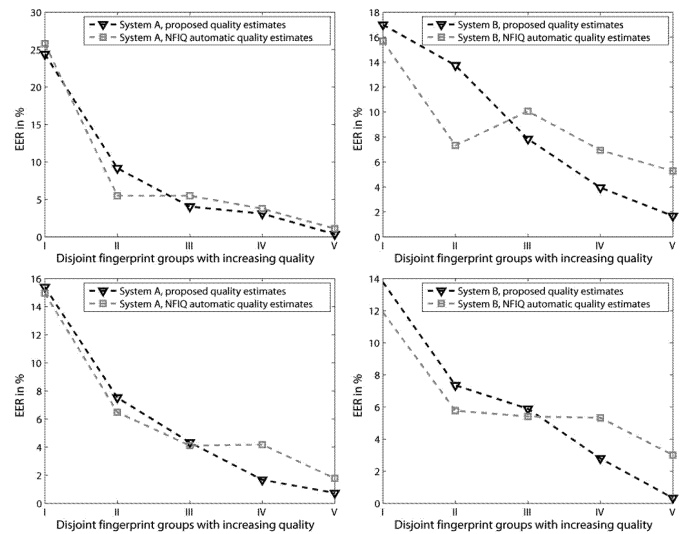

Fig. 8. EER for systems A and B (from left to right) within quality groups I-V from DB2 (top row) and DB3 (bottom row) of FVC2004. Two automatic quality assessment methods are used to establish the partitions (see legend).

In the remaining parts of this section, the three systems A-C are combined (at least two experts at a time) using the fusion schemes explained in Section III. A jackknife (leave-one-out) strategy is employed whenever training is involved, meaning that the training set consists of all users but one (who, together with the impostors, forms the test set), and all users are tested on some point, giving an averaged EER rate. A number of four impressions is used for client and impostor supervisor training, whereas 9, respectively, 74 impressions not belonging to the training set are being tested. Note that each fingerprint is effectively treated as a user and that we take impostors of the same finger type only. When employing nontrained fusion schemes, the test set comprises all users at once, giving 750 × 9 genuine and 750 × 74 impostor trials again.

TABLE I
EER OF SINGLE EXPERTS AND SIMPLE FUSION SCHEMES (MAX/SUM)

|        | A    | B   | C    |     | A,B  | A,C  | B,C  | A,B,C |
|--------|------|-----|------|-----|------|------|------|-------|
| EER %  | 1.22 | 1.9 | 6.37 | SUM | 1.06 | 1.22 | 1.36 | 1.56  |
|        |      |     |      | MAX | 0.75 | 0.84 | 1.16 | 1.16  |

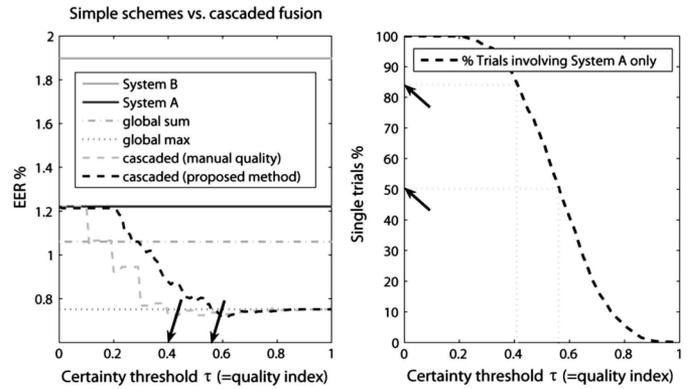

Fig. 9. Left: Results for cascaded fusion compared to simple schemes; The arrows indicate favorable threshold values for the certainty (= image quality) at which the secondary system is triggered. Right: "Efficiency impact" of cascaded fusion, The two arrows are connected to the former arrows through the dotted lines.

The performance (EER) of expert combinations using simple, nonadaptive schemes is given in Table I. We can observe that combinations involving the best expert (system A) deliver the best results, actually outperforming the best expert almost every time. In this test, fusion applying the MAX rule is superior to using SUM, although the former was favored by shifting the experts to a common operating point. The overall best result using simple schemes involves the first two systems and enables a drop in EER of $\approx 38\%$ with respect to the best expert's performance in isolation. It is worth noting that combining all three experts can worsen the joint performance in comparison to selecting only two of them (which need not even be the leading ones). This lies with "simply" fusing experts, which are severely differently skilled, without training.

The left-hand side in Fig. 9 shows the performance of cascaded fusion of systems A and B as a function of certainty $\tau$, chosen as the thresholded quality index. Manual quality estimates are taken in case of the dotted gray line to illustrate a best case, while estimates by our method are considered along the path of the dotted black line. Recognition performance of the single systems, furthermore fused by simple schemes—independent of quality though—are indicated as well, with the MAX rule giving the best result (EER of 0.75%, compare Table I). Employing a cascade with systems A and B as the primary and secondary system, respectively, the 0.75% line is approached from above with a small remainder, considering higher and higher trigger thresholds (image quality). A first minimum, with a difference in EER of 0/0.11% when employing manual/automatic quality indices, respectively, is reached at the threshold marked by the leftmost arrow. The big difference is that in $\approx 84\%$ of all trials, only system A is utilized at this threshold, its "efficiency impact" being marked by the corresponding uppermost arrow to the right in Fig. 9. As illustrated, we (almost) maintain the best error rate for simple fusion of the two systems, but actually need to run system B every sixth time only. Another interesting "operating point" is indicated by the second arrow in the left-hand part of Fig. 9, at which the minimum is reached (EER of 0.75%) while both systems are utilized only half of the time. For these experiments, the MAX rule was employed as a cascaded fusion function $f$. The suggested ad-hoc threshold according to Section III-B would be 0.5. Looking at Fig. 9, it lies in between the previously mentioned "operating points," and leads to an EER of 0.75/0.8%. The efficiency at this point is measured to be approximately 72%, which is even above the theoretical value of 50%.

For the Bayesian-based fusion scheme, indices derived from a quality assessment method are assigned to either one of the systems A-C. This is because we wish to quantify the impact of the image quality on the Bayesian supervisor fusion coupled with a certain expert's ability. The remaining two experts are assigned a quality of 1 (normal) for each trial. The best results in terms of EER are shown in Fig. 10. It turned out that system A was most suitable to attach certainties based on image quality, which is indicated by qA instead of A in Fig. 10. Worth noting, we can observe a drop in EER of $\approx 97/95\%$ when adaptively fusing all experts (qA-B-C) compared to system A in isolation. Adaptive fusion is able to significantly increase recognition performance independently of the quality assessment method employed, while the improvement using three experts compared to two is relatively small. Nevertheless, including system C in the nonadaptive Bayesian supervisor fusion (darkest bars in Fig. 10) leads to an EER drop by $\approx 35\%$. This improvement is remarkably better than in case of the simple fusion schemes where the EER even increases when systems A and B are complemented by system C. This is obviously another effect of training. Previous work has shown that the training of these supervisors is satisfied relatively soon (20 out of 75 users [5]).

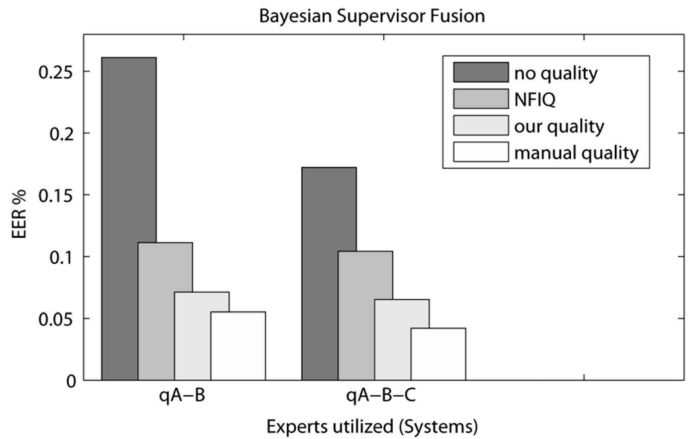

Fig. 10. Best combinations for the Bayesian supervisor fusion. The quality indices were used to weight system A only (therefore qA).

Note that both training and nontraining supervisors are important to different applications as demands on computational efficiencies versus/and decision performance vary. However, in both cases, the automatic quality estimates delivered significant benefits as the experiments indicate. While there have been some studies on how to incorporate quality into training supervisors, the corresponding strategies were largely unstudied for nontraining schemes. The cascade strategy presented before intends to contribute to the latter.

## V. CONCLUSION

We showed how *a priori* content knowledge can be encoded and used in quality estimation. The decomposition of the structure tensor by symmetry features was analyzed for this purpose. Applied to fingerprints, the practical benefit is avoidance of training and adjustment efforts. The experiments show that all fingerprint regions must be treated equally in quality assessment. The proposed method competes well with another, yet heavily trained automatic method (NFIQ) on several databases (verified by the correct quality group division). When exploited to adapt fusion parameters, the levels of agreement studies between human and machine quality assessments have not been reported before, to the best of our knowledge.

We elaborated on the benefits of adapting multialgorithm fusion schemes as a reaction to the signal quality. Experiments with simple schemes (0.75% EER using MAX rule) showed that careless fusion can also increase the EER. As for adaptive fusion, we introduced a nontrained cascaded scheme to dynamically switch on experts in case of uncertainty (low quality), assuming time is the most limited resource. We experimented on two experts in this case, and we could approach the best possible EER, for example, up to a remainder of 0.11% with the help of our automatic quality indices while saving to run the second expert five out of six times. It is also shown for the first time that under certain quality conditions, fusion is expendable. To point out another aspect of multialgorithm fusion, we implemented Bayes-based supervisors for continuous fusion. Taking advantage of training and additionally the quality estimates of the proposed method, (absolute) EERs of 0.17% and 0.07% were achieved, respectively. This was proven by an experiment where quality adaptive fusion and training yield the best recognition rates when combining differently skilled experts.